# A Hybrid Training Algorithm for Continuum Deep Learning Neuro-Skin Neural Network

Mehrdad Shafiei Dizaji, Ph.D.

Department of Engineering Systems and Environment, University of Virginia, VA

*Abstract*— In this brief paper, a learning algorithm is developed for Deep Learning Neuro-Skin Neural Network to improve their learning properties. Neuroskin is a new type of neural network presented recently by the authors. It is comprised of a cellular membrane which has a neuron attached to each cell. The neuron is the cell's nucleus. A neuroskin is modelled using finite elements. Each element of the finite element represents a cell. Each cell's neuron has dendritic fibers which connects it to the nodes of the cell. On the other hand, its axon is connected to the nodes of a number of different neurons. The neuroskin is trained to contract upon receiving an input. The learning takes place during updating iterations using sensitivity analysis. It is shown that while the neuroskin cannot present the desirable response, it improves gradually to the desired level.

*Keywords*— Neuro-skin, Hybrid Algorithm, Particle Swarm Optimization (PSO), L-BFGS-B Algorithm, Training, Neural Networks, Finite Element.

## 1. Introduction

*Neuro-Skins or Nero-Membranes (NMs).* In the previous paper, neuroskins are presented and their response characteristics are studied [1]. Neuroskins are an improved version of Dynamic Plastic Neural Networks (DPNN)s which have recently been introduced by the authors [1-2]. As opposed to a DPCNN which contains a limited number of neurons, each of which is only connected to a small number of points of the base medium, a neuro-skin has the property that every small segment of it has the properties of a neuron. In this regard, the neuro-skin can be considered as a homogeneous skin. This property of the neuro-skin is holographic and does not change with the size of the segment taken out of the membrane. This property will be discussed more in the next sections. The membrane exhibits the activation of neurons. That is at any point it issues an output which is a function of the response of the plate at that point. A more general neuro-membrane can be considered which has the property that its output at any point is a function of the membrane response at farther points as well.

While mathematical modelling and analytical study of neuro-membranes is important to formulate the underlying equations governing their behavior and properties, a first numerical study of their dynamic behavior is essential to provide more insight into their behaviour and to visualize their characteristics. To get more information about the neuroskin neural networks and their characteristics, reader can refer to the previous research [1]. The neuro-membrane utilizing in this paper is a two-dimensional plate which has been studied by the authors in their previous papers on DPCNNs [1] – [26]. They have used this problem in all their studies so that the results and characteristics of the different continuous neural network models could be compared. Figure 1 shows the plate which is 500 mm by 1000 mm. It is unrestricted on three of its edges but restricted by 11 simple supports on its fourth edge [1]. The simple supports provide a large number of redundancies in addition to the reactions needed for the static equilibrium of the

M.S.D. Author is with Department of Engineering Systems and Environment, University of Virginia, VA (phone: 434-987-9780; e-mail: ms4qg@virginia.edu).

plate. The plate is restricted against any rigid-body motion in its plane. Since the problem is studied as plane stress, there is no transverse motion normal to the surface of the plate so that the plate curvature is 0 at any point and in any direction. Also, there is no rotation about any axis within its plane.

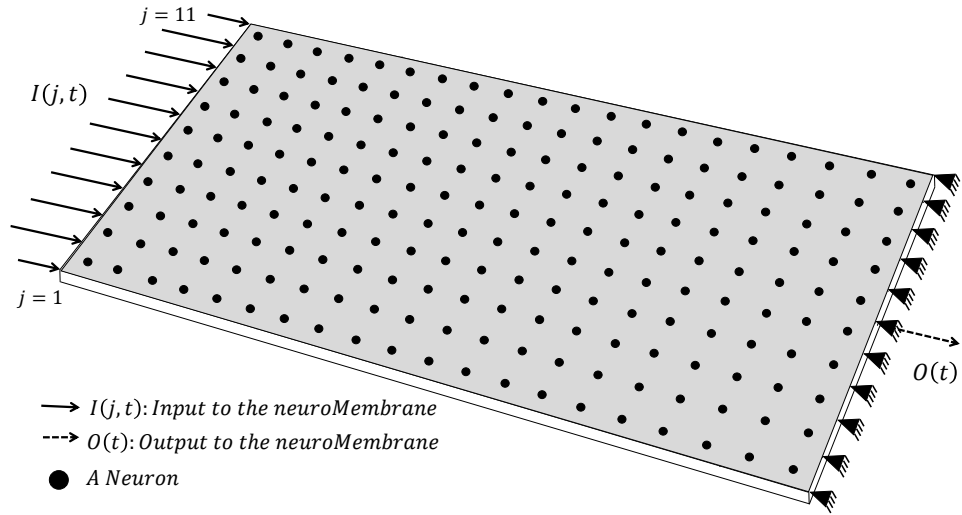

Fig. 1. Plate Neuro-Membrane (NM) studied in this paper with its 11 inputs and its only output. Each big dot represents a neuron. t denotes time and j represents input terminal number [1].

The plate neuro-membrane is modelled by dynamic nonlinear finite element method, and since each element is equipped with a neuron, it is called a Neuro-Element. The discretized membrane itself is then called here Finite-Neuro-Elements (FNEs). Details about the properties of FNEs will be explained in the subsequent sections. The FNEs mesh studied in this paper has 20×10 square neuro-elements, each containing a neuron with an assigned activation function. Fig. 2a shows the membrane when it has been discretized by a mesh of 20×10 FNEs. Fig. 2b shows the order of numbering the nodes.

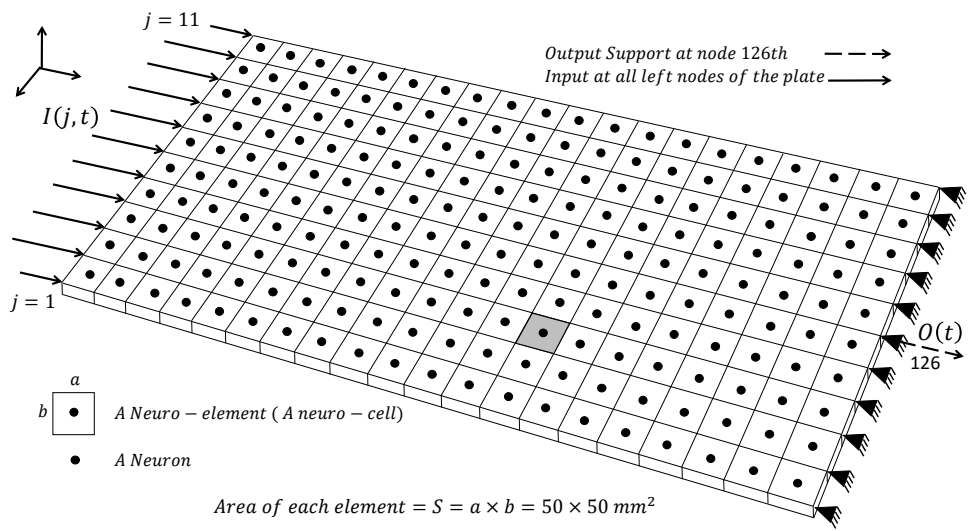

(a)

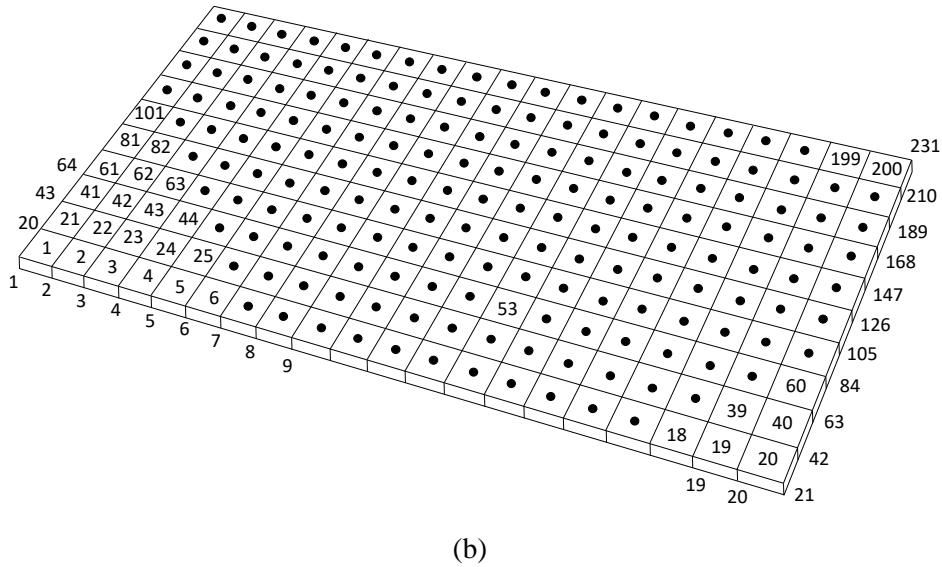

(b)

Fig. 2. Finite-Neuro-Elements mesh studied in this paper has 10×20 square Neuro-Elements (cells), each containing a neuron. The size of all the square elements is 50 mm: (a) inputs, output and elements with their neurons, (b) how the nodes and elements have been numbered. Node number 126 is the output node of the neuro-membrane [1].

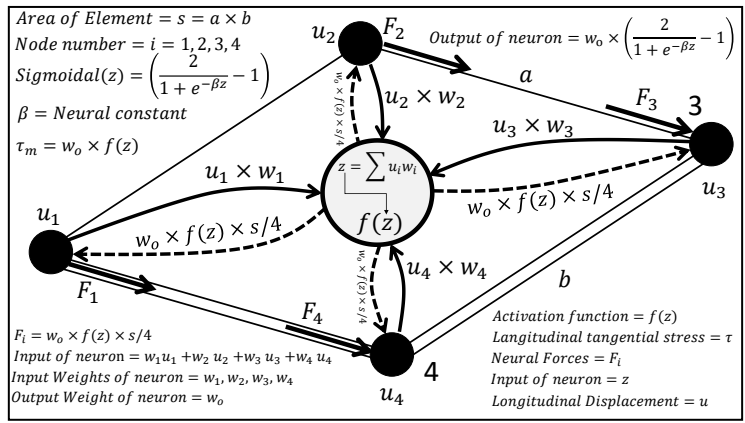

Fig. 3. Isometric view of a 4-node rectangular Neuro-Element with its neuron. For the square element used in this paper, $a=b$. The neuron shown here is sigmoidal but can be any type neuron, as is discussed later in the paper [1].

Figure 3 shows an isometric view of a neuro-element used in this paper, with all the information about its neuron, input connection weights and its output weight. The figure also explains how the output of the neuro-element is calculated. The horizontal displacements of nodes are multiplied by their corresponding input weights. The result is delivered to the neuron's processor where $f(z)$, the neuron's activation function, receives the input z and issues its output $f(z)$ which is a value in [-1, +1]. The output is first multiplied by the neuron's output weight ($w_o$), then multiplied by the neuro-element's area to determine the total traction force of the element. The traction is divided by 4 and is applied to each node of the neuro-element to be assembled later as a part of the total nodal force. The behaviour and characteristics of a neuro-membrane depends on the type of the activation function incorporated in it. This issue is studied in the following sections where different types of activation functions are considered for the example problem in this paper and the response of the neuro-membrane has been studied. Also, the adaptivity of the neuro-membranes to learn a dynamic input-output relationship, similar to multi layer feed forward neural networks, is investigated briefly without aiming to provide final details [1].

## 2. Training Algorithm Neuro-Skins or Nero-Membranes (NMs)

*A Hybrid Optimization Algorithm.* The adaptive parameters of the Neuro-Skin neural network are its connection weights. Generally, the error of the Neuro-Skin can be expressed in terms of the adaptive parameters. The mean Squared Error (MSE) of the Neuro-Skin Neural Network is:

$$MSE(w) = \frac{1}{N}\sum_{i=1}^{N}(\hat{y}(t_i) - y(t_i))^2 \qquad (1)$$

Where $w$, $N$ and $\hat{y}(t_i)$ are connection weights, number of training pairs and output of the Neuro-Skin Neural Network, respectively.

*Particle Swarm Optimization (PSO).* In this proposal the Particle Swarm Optimization (PSO) algorithm is adopted to find a local optimal solution for the above-mentioned problem. Then the L-BFGS-B method is successfully applied for training process of the Neuro-Skin Neural Network. In the Figure 4, training procedure of the Neuro-Skin Neural Network is shown schematically. The design variables are selected as an output to each neuron which are tuned during learning procedure. The results for convergence of the design variables to their target values indicates the capability of the introduced training algorithm. The original PSO suggested by Kennedy and Eberhart is based on the analogy of swarm of bird and school of fish. Each particle in PSO makes its decision using its own experience and its neighbor's experiences for evolution. In this problem, the cost function is defined as follows:

$$\min\ J_p = \sum_{t=0}^{T}\sum_{i=1}^{n}\left\|y_{ti_p}^{FEM} - y_{ti}^{DIC}\right\|^2,\ \forall p \in \{1,2,\ldots,P\} \qquad (2)$$

Where, $T$ is the simulation time (second), and $n$ is the number of defined nodes, $p$ shows the particle and $P$ is the total number of particles in each iteration. $y_{ti_p}^{FEM}$ and $y_{ti}^{DIC}$ are the measurements (i.e., displacement and strain) obtained from simulation from particle $p$ and DIC at time step $t$ for node $i$, respectively. The solution vector that are generated and updated by PSO algorithm for particle $p$ in each iteration is $E_p = [e_p^1, e_p^2, \ldots, e_p^n]$ that is constitutive properties of material of finite elements. The architecture of the PSO algorithm will be designed to come up with the best guess of $E$ vector such that the difference between the average displacement of the FEM and DIC through the simulation time to be in its minimum possible amount.

The Particle Swarm Optimization algorithm is one of the recently proposed heuristics (metaheuristic) algorithms. The idea behind the evolutionary algorithms is that the mathematical model for the system is not available or its model is such complicated that dealing with that is computationally expensive. The PSO algorithm can be applied to nonlinear non-convex and non-continuous optimization problems. It is a population-based search algorithm and starts with a group of particles (i.e. defined solution $E$). These particles or solutions search in the solution space from one point to another by using the last solution and a velocity vector (Equation 3).

$$E_p^{k+1} = E_p^k + V_p^k \qquad (3)$$

In this equation, $k$ and $p$ show the number of iteration and particle, and $V_p^k$ is the velocity vector of the particle $p$ at iteration $k$. In each iteration, after calculating the fitness value for each particle we save their minimum and compare it with the previous step elite and save it as the global minimum till now and show it by $E_g$. The algorithm keeps doing this comparison and saving process by the end of algorithm's iterations and return it as the best-found solution. Also, the algorithm records the best solution found by each particle by the current iteration. This solution is represented by $E_{bp}$. The velocity vector can be defined as follow (Equation 4):

$$V_p^k = \omega V_p^{k-1} + \varphi_1 R(E_{bp}^k - E_p^k) + \varphi_2 C(E_g^k - E_p^k) \qquad (4)$$

Where, $\omega$ is inertial weight, $\varphi_1$ and $\varphi_2$ are called acceleration coefficients. $R$ and $C$ are two $n \times n$ diagonal matrices in which the entries in the main diagonal are random numbers uniformly distributed in the interval $[0,1)$. At each iteration, these matrices are regenerated. By adjusting these factors, we can change the rate of convergence, running time and exploitation/exploration proportion. The main advantages of the PSO algorithm are summarized as: simple concept, easy implementation, and computational efficiency when compared with mathematical algorithm and other heuristic optimization techniques such as Genetic Algorithm [68]. Figure 5 represents a conceptual scheme for the PSO algorithm's steps.

For the demonstration purposes, the Particle Swarm Optimization algorithm is utilized for a rectangular 2D numerical model with only four design parameters. Figure 4 shows the finite element numerical model along with the results after optimization process. The initial and updated values for the design parameters which are $E_1, E_2, E_3, E_4$ are shown in the Figure 4. Also, the target values for the design parameters are selected to be 500000. As it can be seen in the Figure 4, the Hybrid Particle Swarm Optimization was able to modify the design parameters appropriately so that they could reach to their target value after the proposed number of iterations accordingly. This demonstration example only had four different design parameters. However, in a complicated 3D structural element, there will be large number of design parameters. Hence, the necessity of developing a multi-processing optimization is inevitable. To increase the efficiency and effectiveness of the optimization algorithm, a multi-processing algorithm is introduced. For any number of $x$ design parameters in the code, $x$ number of CPU is allocated so that all the decision parameters are updated simultaneously. In the Figure 4 this concept is shown schematically.

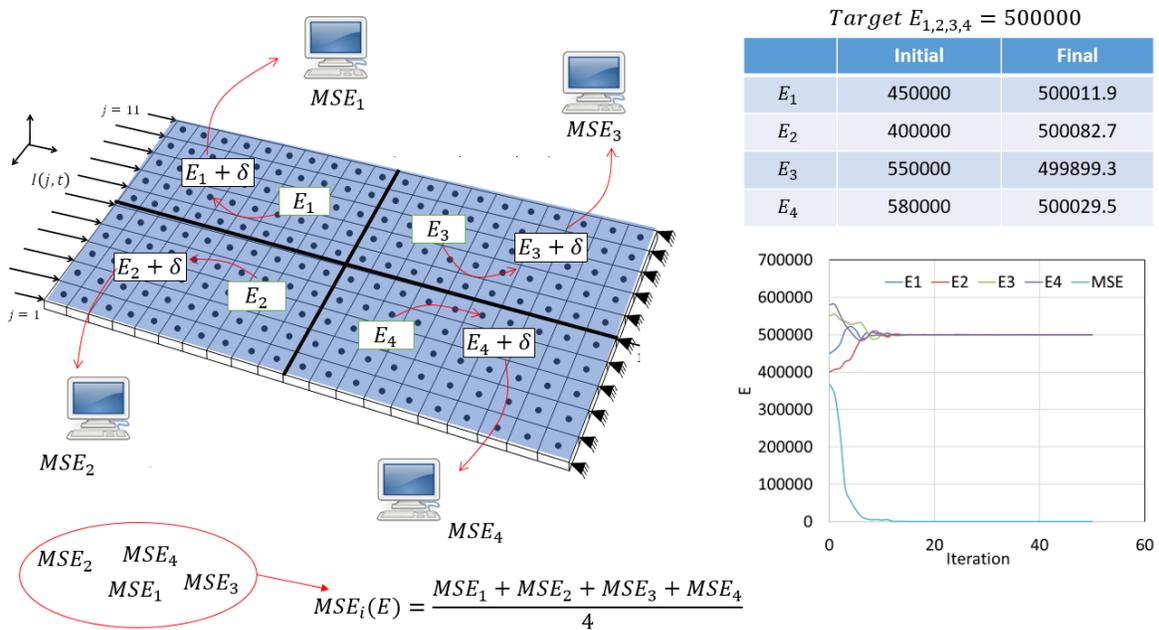

Fig. 4. Training algorithm of the Deep Learning Neuro-Skin Neural Network schematically

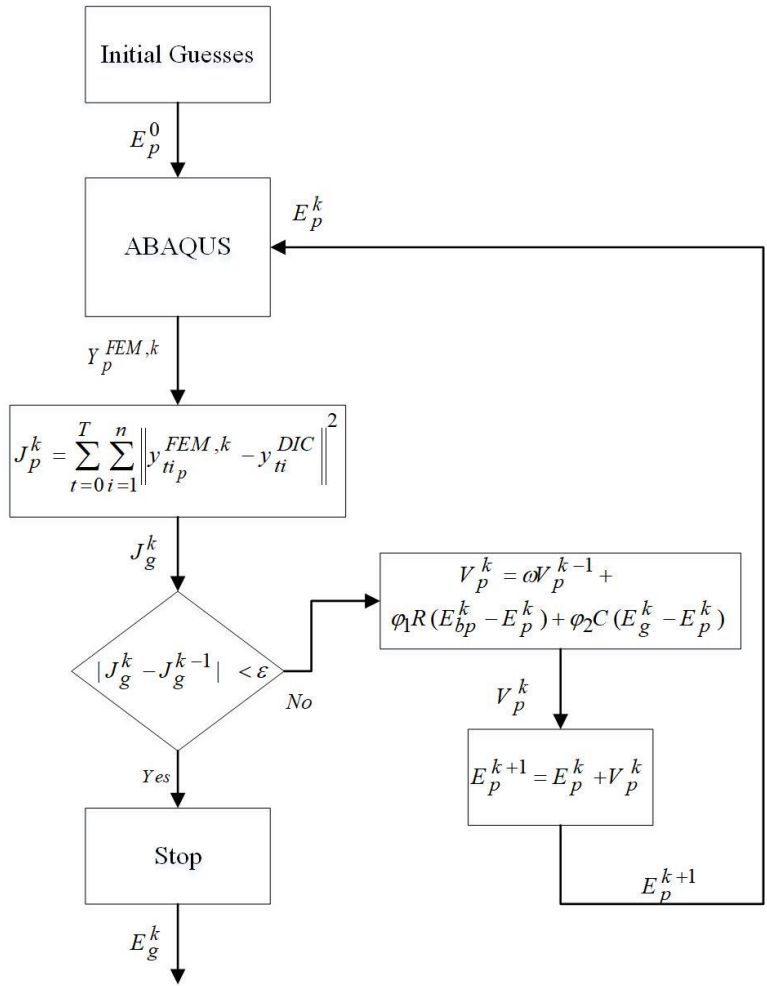

Fig. 5. The flowchart showing Particle Swarm Optimization

## 3. Conclusion

A new algorithm based on sensitivity analysis for the process of training of Neuro-skin was introduced. The neuro-skin is adaptive and has the capability of providing desirable response to inputs intelligently. Hence, after training, it responds as a smart medium and this way, the neuro-skin can be considered as a new type of neural network, called Deep Learning Neuro Skin Neural Networks, with adaptability and learning capability. Training of the neuro-skin to represent a desired output was an essential part of the study. The paper shows that the defined training algorithm of the neuro-skin by dynamic finite element method is a suitable approach.

**Appendix**

The programming code is provided as follows:

```python
from multiprocessing import Pool,cpu_count
import subprocess
import os
import shutil
import fileinput
import numpy as np
from scipy.optimize import fmin_bfgs,fmin,fmin_l_bfgs_b

ansys_dir='C:\\Program Files\\ANSYS Inc\\v171\\ansys\\bin\\winx64'

target=np.loadtxt('output.out')

def run(num,x):
    if not os.path.exists(str(int(num))):
        os.makedirs(str(int(num)))
        current_dir=os.getcwd()
        current_dir +='\\'+str(int(num))
        shutil.copy('aaa.txt',current_dir)
        shutil.copy('bbb.txt',current_dir)
        shutil.copy('fne.txt',current_dir)
    else:
        current_dir=os.getcwd()
        current_dir +='\\'+str(int(num))

    with open(current_dir+'\\aaa.txt','w') as f:
        k=200/len(x)
        for i in range(len(x)):
            for j in range(k):
                m=1+i*k+j
                f.write('v'+str(int(m))+'='+str(x[i])+'\n')

    command=ansys_dir+'\\ansys171 -b -np 1 -dir "'+current_dir +'" -i fne.txt -o analysis.o'
    os.chdir(current_dir)
    subprocess.call(command)
    return np.loadtxt('output.out')

def run_ex(x):
    return run(*x)

def func(x):
    p=Pool(len(x)+1)
    input_list = [(0,x)]
    delta = 1.0e-2
    for i in range(len(x)):
        copy = np.array(x)
        copy[i] += delta
        input_list.append((i+1,copy))
    out=p.map(run_ex,input_list)
    f=np.sqrt(np.mean((out[0]-target)**2))
    g=[]
    for i in range(len(x)):
        f1=np.sqrt(np.mean((out[i+1]-target)**2))
        g.append((f1-f)/delta)
    return f,np.array([g])

def evaluate(xx):
    p=Pool(cpu_count())
    input_list=[]
    for i in range(len(xx)):
        input_list.append([i,xx[i]])
    out=p.map(run_ex,input_list)
    f=[]
    for i in range(len(out)):
        f.append(np.sqrt(np.mean((out[i]-target)**2)))
    return f

def call_back(xk):
    with open('result.txt','a') as f:
        st=''
        for xx in xk: st+=str(xx)+','
        st=st[:-1]+'\n'
```

```python
            f.write(st)

if __name__ == '__main__':
    if os.path.isfile('result.txt'):
        os.remove('result.txt')
    xopt,fopt,d= fmin_l_bfgs_b(func, x0=[450000.0],bounds=[[400000,550000]],approx_grad=False,epsilon=0.001,factr=1.0e12,maxfun=100,maxiter=5,iprint=1,callback=call_back)
    #f=open("final.txt",'w')
    #f.write(str(xopt)+','+str(fopt))
    #f.close()
    print (xopt,fopt)
    x=np.loadtxt('result.txt',delimiter=',')
    x=np.reshape(x,(len(x),-1))
    x=x.tolist()
    mse=evaluate(x)
    out=np.column_stack((x,mse))
    np.savetxt('result.txt',out,delimiter=',')
```